\documentclass{llncs}

\usepackage[utf8x]{inputenc} % allow utf-8 input
\usepackage[T1]{fontenc}    % use 8-bit T1 fonts
\usepackage{textcomp}
\usepackage{dsfont}
\usepackage{calligra}
\usepackage{setspace}
\usepackage{mathrsfs}
\usepackage{ragged2e}
\usepackage{url}            % simple URL typesetting
\usepackage{booktabs}       % professional-quality tables
\usepackage{amsfonts}       % blackboard math symbols
\usepackage{nicefrac}       % compact symbols for 1/2, etc.
\usepackage{microtype}      % microtypography
\usepackage{xcolor}         % colors

\DeclareUnicodeCharacter{2009}{\,}
\DeclareUnicodeCharacter{00D6}{\"O}
\UseRawInputEncoding

\usepackage[noadjust]{cite}

\usepackage{xcolor,soul,framed} %,caption

\colorlet{shadecolor}{yellow}
\usepackage[pdftex]{graphicx}
\graphicspath{{../pdf/}{../jpeg/}}
\DeclareGraphicsExtensions{.pdf,.jpeg,.png}

\usepackage{array}
\usepackage{mdwmath}
\usepackage{mdwtab}
\usepackage{eqparbox}
\usepackage{url}
\usepackage{multirow}
\usepackage{graphicx}
\usepackage{amsfonts}
\usepackage{amsmath}
\usepackage{amssymb}
\usepackage{graphicx}
\usepackage{algorithm}
\usepackage{algpseudocode}
\usepackage{multicol}
\usepackage{multirow}
\usepackage{adjustbox}
\usepackage[colorlinks,urlcolor=purple,linkcolor=purple,citecolor=purple]{hyperref}
\usepackage{bm}
\usepackage{float}
\usepackage{booktabs}
\usepackage{longtable}
\usepackage{diagbox}

\usepackage[caption=false, font=footnotesize]{subfig}
\definecolor{lightblue}{rgb}{0.9, 0.95, 1.0}
\definecolor{mygreen}{rgb}{0.01, 0.5, 0.01}
\definecolor{myred}{rgb}{0.8, 0.01, 0.01}
\definecolor{customgray}{rgb}{0.25,0.25,0.25}
\definecolor{customred}{rgb}{0.8,0.05,0.05}

\usepackage{pifont}
\usepackage{url}
\usepackage{adjustbox}
\usepackage{pdflscape}
\usepackage{anyfontsize}
\newcommand{\repeatthanks}{\textsuperscript{\thefootnote}}
\usepackage{soul}
\usepackage[moderate,tracking=normal,paragraphs=normal]{savetrees} % paragraphs=normal,tracking=normal,

\begin{document}

\title{Vision Transformers for End-to-End Quark-Gluon Jet Classification from Calorimeter Images}
\titlerunning{Vision Transformers for End-to-End Quark-Gluon Jet Classification}

% \author{Anonymous Author(s)\inst{1}}
% \institute{Affiliation\\Address\\
% \email{email}
% }

\author{Md Abrar Jahin\inst{1}\thanks{Corresponding Author(s)}\orcidID{0000-0002-1623-3859}
\and Shahriar Soudeep\inst{2}\orcidID{0009-0004-9317-2326}
\and Arian Rahman Aditta\inst{3}
\and M. F. Mridha\inst{2}\repeatthanks\orcidID{0000-0001-5738-1631}
\and Nafiz Fahad\inst{4}
\and Md. Jakir Hossen\inst{4}\repeatthanks\orcidID{0000-0002-9978-7987}
}

\authorrunning{Md Abrar Jahin et al.}
\institute{University of Southern California\\
\email{jahin@usc.edu}
\and American International University-Bangladesh\\
\email{20-43823-2@student.aiub.edu, firoz.mridha@aiub.edu}
\and Khulna University of Engineering \& Technology\\
\email{aditta2009028@stud.kuet.ac.bd}
\and Multimedia University\\
\email{fahadnafiz1@gmail.com, jakir.hossen@mmu.edu.my}
}

\maketitle

\begin{abstract}
Distinguishing between quark- and gluon-initiated jets is a critical and challenging task in high-energy physics, pivotal for improving new physics searches and precision measurements at the Large Hadron Collider. While deep learning, particularly Convolutional Neural Networks (CNNs), has advanced jet tagging using image-based representations, the potential of Vision Transformer (ViT) architectures, renowned for modeling global contextual information, remains largely underexplored for direct calorimeter image analysis, especially under realistic detector and pileup conditions. This paper presents a systematic evaluation of ViTs and ViT-CNN hybrid models for quark-gluon jet classification using simulated \textit{2012 CMS Open Data}. We construct multi-channel jet-view images from detector-level energy deposits (ECAL, HCAL) and reconstructed tracks, enabling an end-to-end learning approach. Our comprehensive benchmarking demonstrates that ViT-based models, notably \textit{ViT+MaxViT} and \textit{ViT+ConvNeXt} hybrids, consistently outperform established CNN baselines in F1-score, ROC-AUC, and accuracy, highlighting the advantage of capturing long-range spatial correlations within jet substructure. This work establishes the first systematic framework and robust performance baselines for applying ViT architectures to calorimeter image-based jet classification using public collider data, alongside a structured dataset suitable for further deep learning research in this domain. The implementation of our code is available at: \url{https://github.com/Abrar2652/particle_reconstruction/}.

\end{abstract}

% === I. INTRODUCTION =============================================================
% =================================================================================
\section{Introduction}
The analysis of hadronic jets, collimated sprays of particles produced by the hadronization of quarks and gluons, is central to precision measurements and new physics searches at the CERN Large Hadron Collider (LHC)~\cite{larkoski_jet_2020}. Among the various jet classification problems, distinguishing between quark- and gluon-initiated jets remains particularly important and challenging. Many signatures of Standard Model (SM) processes and beyond-the-SM scenarios involve final states with quark or gluon jets, and accurate identification of their origin can directly improve background rejection and signal sensitivity in collider analyses~\cite{larkoski_jet_2020}. Quark-gluon discrimination is difficult because both types of jets emerge from QCD interactions and share overlapping kinematic and topological features. The difference lies in subtle variations in their radiation patterns, color factors, and particle multiplicities. Gluon jets typically produce higher particle multiplicity, broader radiation patterns, and softer transverse momentum spectra due to their larger color charge~\cite{larkoski_jet_2020}. Capturing these subtle differences requires classifiers capable of learning fine-grained spatial and structural patterns in the detector data, often in the presence of pileup contamination and detector noise.

Recent advances in machine learning (ML), particularly deep learning (DL), have transformed jet tagging by enabling algorithms to learn directly from low-level detector or reconstructed particle data, bypassing hand-engineered observables~\cite{qu_jet_2020}. At the CMS experiment, conventional approaches for particle and jet identification have historically relied on complex reconstruction algorithms that combine information from various sub-detectors~\cite{Sirunyan_2017}. However, recent studies have demonstrated that applying convolutional neural networks (CNNs) directly to detector-level images can yield competitive classification performance, without relying on the full reconstruction chain, through an end-to-end learning paradigm~\cite{andrews_end--end_event,andrews_end--end_jet}.

While CNNs operating on jet images and particle cloud networks have achieved remarkable success, they often rely on local receptive fields and predefined hierarchical structures, limiting their ability to capture long-range correlations within a jet's spatial energy deposit profile. These global contextual dependencies can be crucial for separating quark and gluon jets, especially in high-pileup environments. In contrast, transformer-based architectures, which have demonstrated state-of-the-art performance in natural language processing~\cite{vaswani_attention_2017} and computer vision~\cite{dosovitskiy_image_2020,liu_swin_2021}, offer a compelling alternative. Vision Transformers (ViTs)~\cite{dosovitskiy_image_2020} and their hierarchical variants, such as Swin Transformer~\cite{liu_swin_2021} and MaxViT~\cite{avidan_maxvit_2022}, have shown strong capabilities in modeling global image dependencies without relying on inductive biases inherent to CNNs. Their attention-based mechanism allows the model to dynamically focus on important spatial regions in the input, which is highly desirable for interpreting sparse and irregular energy patterns in jet images. Despite their success in other domains, the application of ViT-based models to calorimeter image data for jet classification remains largely unexplored. Existing studies in high-energy physics (HEP) predominantly focus on particle cloud representations~\cite{qu_jet_2020,guo_pct_2021}, leaving a gap in evaluating transformer-based models directly on image-based representations of jet substructure. Furthermore, publicly available, realistic datasets like the \textit{CMS Open Data} provide an ideal testbed to assess these models under conditions closely matching those of operational collider experiments.

In this study, we present a systematic evaluation of ViTs and ViT-CNN hybrid architectures for quark-gluon jet classification using simulated \textit{2012 CMS Open Data}~\cite{andrews_end--end_jet}. We construct jet-view images incorporating energy deposits from the ECAL, HCAL, and reconstructed tracking systems, forming multi-channel images that preserve the spatial distribution of detector signals within a fixed angular region. This end-to-end approach, free from particle reconstruction algorithms, provides the classifier direct access to rich detector-level data in all its complexity~\cite{andrews_end--end_event,andrews_end--end_jet}, offering an opportunity to test ViTs as an alternative to CNNs for calorimeter image analysis at the LHC. Our work benchmarks the performance of ViT-based models against competitive CNN baselines and investigates their sensitivity to key training hyperparameters through an extensive ablation analysis.

Our \textbf{key contributions} are as follows: \textbf{(1)} We propose the first systematic framework applying ViTs and ViT-CNN hybrid models to calorimeter image-based quark-gluon jet classification using CMS Open Data.
\textbf{(2)} We construct a structured calorimeter image dataset from simulated \textit{2012 CMS Open Data}, incorporating realistic detector and pileup conditions, suitable for evaluating deep learning models for jet classification tasks.
\textbf{(3)} We conduct an extensive performance evaluation and hyperparameter sensitivity analysis of ViT-based and CNN-based models, establishing new baselines for transformer architectures in end-to-end detector image-based jet classification tasks.

% While deep learning-based jet tagging has matured considerably through particle cloud and image-based CNN models, transformer architectures remain underutilized in this domain. Existing models like ParticleNet dominate cloud-based approaches, but few studies have evaluated the efficacy of Vision Transformers for calorimeter image data. Given the transformer's capacity to model long-range spatial dependencies and the success of ViT variants in dense visual recognition tasks, this work investigates their potential for quark-gluon jet classification using CMS Open Data. In doing so, we aim to bridge the methodological gap between recent breakthroughs in computer vision and collider physics analysis.

% The paper is structured as follows: Section II reviews related work, while Section III describes our dataset, the proposed QRGCL model, and the benchmark models. Section IV discusses the experimental setup, and Section V presents the results and discussion. Finally, Section VI summarizes the key findings and suggests directions for future research.

\section{Related Work}
\label{related_work}

\subsection{Deep Learning for Jet Tagging}
DL has transformed jet tagging at the LHC, achieving unprecedented accuracy by leveraging various data representations. Early approaches included 2D CNNs for calorimeter images~\cite{de_oliveira_jet-images_2016}, sequential models for particle lists~\cite{guest_jet_2016}, recursive networks~\cite{louppe_qcd-aware_2019}, and graph neural networks~\cite{henrion_neural_2017}. The field has since shifted toward particle cloud representations inspired by computer vision point cloud methods~\cite{qu_jet_2020}, notably through the Energy Flow Network (EFN)\cite{komiske_energy_2019} and ParticleNet\cite{qu_jet_2020}, both built on the Dynamic Graph CNN (DGCNN) framework~\cite{wang_dynamic_2019}. While subsequent works explored GAPNet-like~\cite{chen_fair_2022} and transformer-based point cloud models~\cite{mikuni_point_2021}, none have decisively surpassed ParticleNet on public benchmarks. Parallel efforts introduced physically motivated inductive biases, including the Lund jet plane~\cite{dreyer_leveraging_2022}, Lorentz-equivariant networks~\cite{jahin_lorentz-equivariant_2025,gong_efficient_2022}, and rotationally invariant models~\cite{dillon_symmetries_2022}. These advances improved robustness, data efficiency, and interpretability, with a notable impact on LHC analyses. The CMS DeepAK8 algorithm~\cite{sirunyan_identification_2020} and ParticleNet have enabled key results, including the first observation of $Z \rightarrow c\bar{c}$ decays~\cite{tumasyan_search_2023} and rare Higgs process searches~\cite{cms_collaboration_search_2023}.

\subsection{Transformer Architectures in Scientific Applications}
Transformers, introduced in NLP by Vaswani et al.~\cite{vaswani_attention_2017} and extended to vision via ViT~\cite{dosovitskiy_image_2020} and Swin Transformer~\cite{liu_swin_2021}, have redefined state-of-the-art models across ML. Landmark applications in the sciences include AlphaFold2~\cite{jumper_highly_2021}, which embedded relational biases into attention mechanisms for protein structure prediction. In HEP, transformer adoption remains at an early stage. JetBERT~\cite{mikuni2025} applied BERT-style models to particle sequences, while CaloFlow~\cite{andrews_end--end_jet} introduced attention to normalizing flows for calorimeter simulation. Some recent works explored transformer variants for jet tagging, focusing mainly on particle cloud representations or ViT-style models for image-based data. However, systematic studies benchmarking ViTs on realistic, pileup-affected calorimeter data--where attention's global context modeling could be advantageous--are lacking~\cite{furuichi_jet_2024}.

\section{Methodology}
\label{methodology}
This study evaluates ViT-based models, alongside established CNN architectures and hybrid approaches, for classifying quark and gluon jets in particle collision data. The methodology consists of several stages: \textbf{first}, we describe the quark-gluon dataset, including its origin, structure, and an 80:20 train/validation split; \textbf{second}, we apply a multi-stage preprocessing pipeline involving zero suppression, channel-wise Z-score normalization, channel-wise max value clipping, and sample-wise min-max scaling, followed by common augmentation techniques such as random horizontal flipping, rotations, resized cropping, and color jittering; \textbf{third}, we evaluate various model architectures, including ViT variants (e.g., ViT-Base), hierarchical transformers (e.g., Swin Transformer, MaxViT), CNN-based models (e.g., ResNet, EfficientNet, ConvNeXt, RegNetY, CoAtNet), and hybrid models combining transformer and CNN features; \textbf{fourth}, we detail the training procedure, including optimization strategies (e.g., AdamW), learning rate scheduling (e.g., Cosine Annealing), mixed-precision training, and gradual unfreezing of pre-trained models; \textbf{finally}, we evaluate model performance using accuracy, precision, recall, F1 score, and ROC-AUC, along with analysis of training time, inference speed, and parameter counts. This approach facilitates a thorough comparative analysis of the models' ability to classify quark and gluon jets.

\subsection{Dataset Description}
In this study, we employ the publicly available \textit{2012 CMS Open Data}\footnote{\url{https://opendata.cern.ch/docs/cms-getting-started-2011}}, released by the CMS Collaboration at CERN. This dataset provides a high-fidelity simulated sample of proton-proton collision events at a center-of-mass energy of 8 TeV, using detailed detector simulations based on the \texttt{Geant4} toolkit. \texttt{Geant4}\footnote{\url{https://geant4.web.cern.ch}} offers state-of-the-art, first-principles modeling of particle interactions with matter, as well as the most complete geometrical description of the CMS detector systems.

\begin{figure}[!ht]
    \centering
    \subfloat[Single event visualization\label{fig:single_event}]{
       \includegraphics[width=0.43\linewidth]{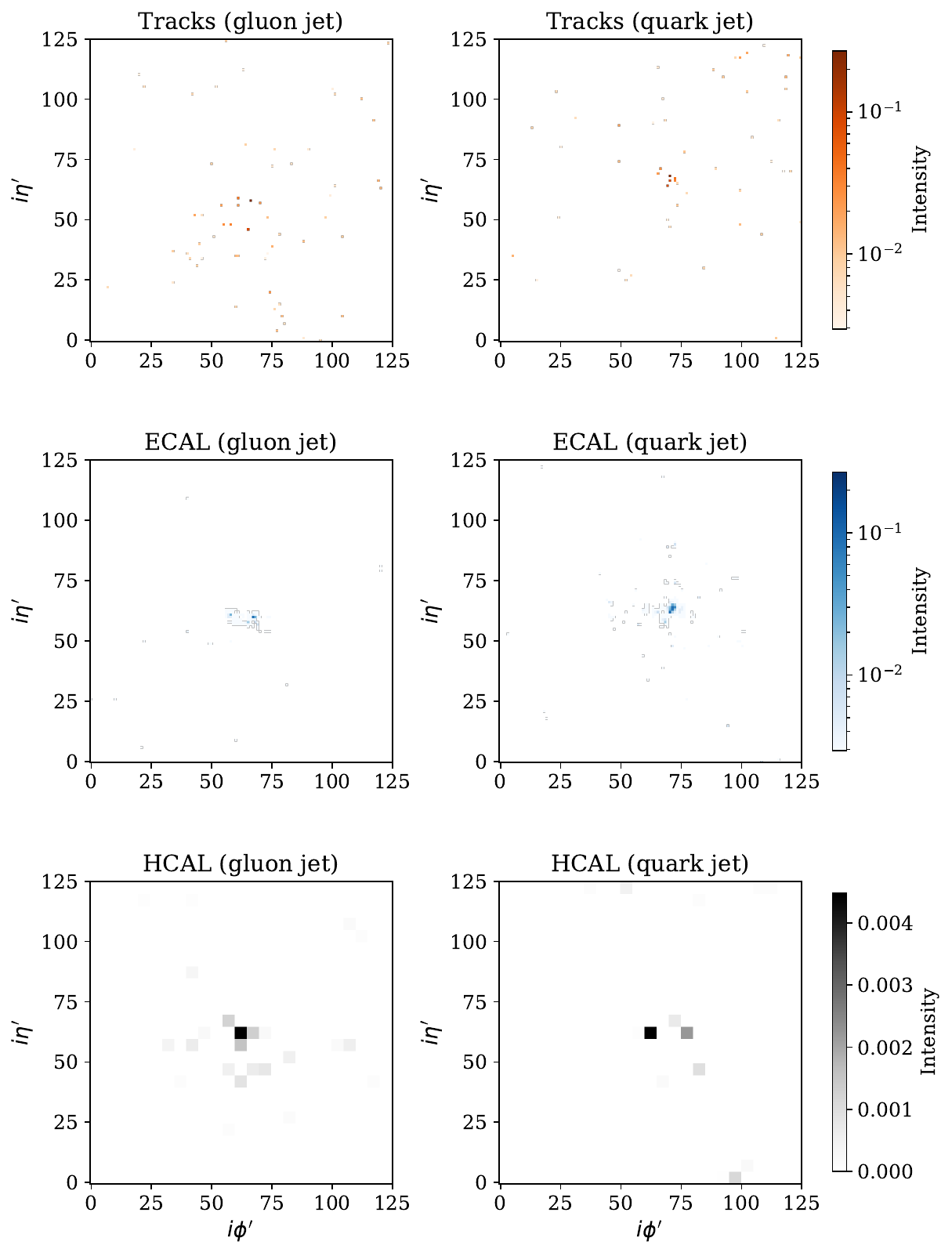}}
    \hfill
    \subfloat[Average per-pixel intensity distributions\label{fig:average_intensity}]{
        \includegraphics[width=0.47\linewidth]{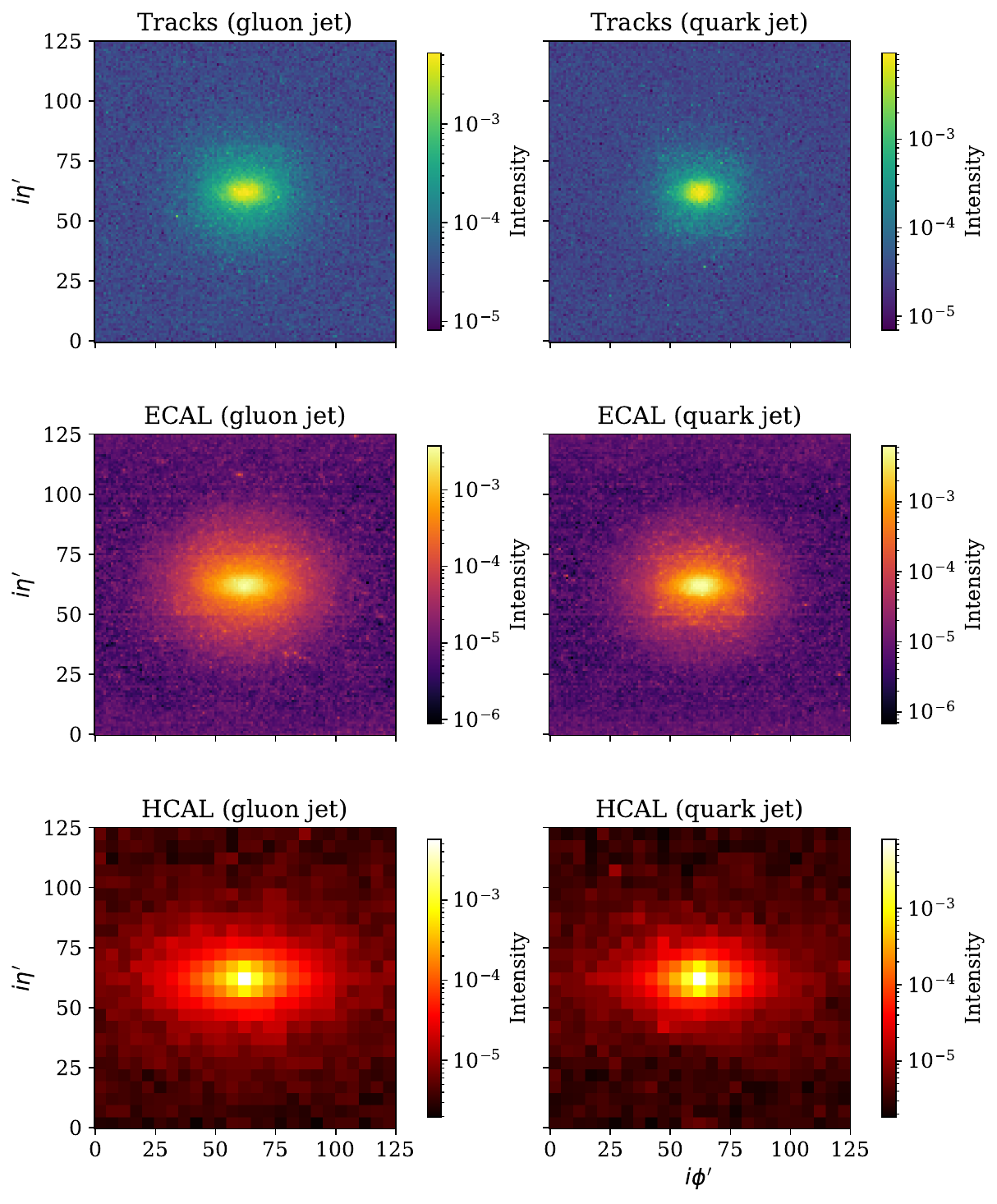}}
    \caption{\textbf{(a)} Visualization of a representative gluon \textbf{(left)} and quark \textbf{(right)} jet event across the Tracks, ECAL, and HCAL detector channels. Logarithmic normalization is used for Tracks and ECAL, while linear scaling is applied to the sparse HCAL data. Colorbars indicate per-channel pixel intensities. \textbf{(b)} Average per-pixel intensity maps over $N = 10^4$ gluon \textbf{(left)} and quark \textbf{(right)} jets for each detector channel. Logarithmic scaling highlights the dynamic range of intensities; colorbars denote per-channel values.}
    \label{fig:enter-label}
\end{figure}

\subsubsection{Simulated Data Samples and Event Selection}
For our quark-gluon jet classification study, we utilize simulated events from QCD dijet production processes, where the generated parton-level transverse momentum ($\hat{p}_T$) lies within the range of 90 to 170~GeV~\cite{cms_collaboration_gjet_pt40_doubleemenriched_tunez2star_8tev_ext-pythia6_2017}. The events were generated and hadronized using the \textsc{Pythia~6} Monte Carlo event generator with the Z2$^*$ tune, which captures the characteristic differences in hadronization patterns between quark- and gluon-initiated showers. These simulations incorporate realistic multi-parton interactions from the underlying event and replicate run-dependent pileup (PU) conditions consistent with the 2012 LHC data-taking periods, with an average pileup of 18--21 interactions per event. Following event generation, a strict selection is applied to isolate a clean sample suitable for quark-gluon discrimination. Only events in which both outgoing partons from the Pythia hard scatter are either light quarks ($u$, $d$, $s$) or gluons are retained. Each parton is geometrically matched to a reconstructed jet within a cone of $\Delta R < 0.4$ in the pseudorapidity-azimuth ($\eta$-$\phi$) plane, and jets with wide-angle radiation beyond this cone are excluded to simplify the classification task. The selected jets must satisfy kinematic requirements of $p_T > 70$~GeV and $|\eta| < 1.8$. For consistency, only the leading-$p_T$ jet from each passing event is retained.

The dataset used in this work is derived from the open-source jet image dataset described by Andrews et al.~\cite{andrews_end--end_jet}, based on simulated QCD dijet events made available by the CERN CMS Open Data Portal. The dataset consists of 933,206 3-channel images of size $125 \times 125$, with half representing quark jets and the other half gluon jets. Each image encodes the energy deposits of a single jet in three different subdetectors of the CMS experiment: the inner tracking system (Tracks), which identifies charged particle trajectories; the electromagnetic calorimeter (ECAL), which measures energy deposits from electromagnetic particles; and the hadronic calorimeter (HCAL), which records energy from hadronic showers. In the CMS experiment, particle momenta are measured in a right-handed coordinate system centered at the nominal interaction point. The $x$-axis points radially inward toward the LHC ring center, the $y$-axis points vertically upward, and the $z$-axis follows the beam direction. The azimuthal angle $\phi$ is measured in the $x$-$y$ plane from the $x$-axis, and the polar angle $\theta$ is measured from the $z$-axis. For collider analyses, the pseudorapidity $\eta$ is frequently used in place of $\theta$ and is defined as
\begin{equation}
\eta \equiv -\ln \left[ \tan \left(\frac{\theta}{2}\right) \right].
\end{equation}
The transverse momentum $p_T$ of each particle is calculated as
\begin{equation}
p_T \equiv \sqrt{p_x^2 + p_y^2}.
\end{equation}
These jet images are constructed by projecting detector hits onto a two-dimensional grid in the $\eta$-$\phi$ plane centered on the jet axis, with pixel intensities corresponding to the $p_T$-weighted energy deposits in each subdetector.

\subsubsection{CMS Detector Image Construction}
The CMS detector is structured as a series of concentric cylindrical systems, consisting of an inner tracking system, followed by ECAL and HCAL calorimeters, and an outer muon system. The CMS Open Data includes calibrated, reconstructed hits~\cite{collaboration_cms_2008} from the ECAL at the crystal level and from the HCAL at the tower level~\cite{andrews_end--end_event}. Additionally, track information is provided via reconstructed track fit parameters, as direct hit-level data from the tracking detectors is not included in the open dataset.

For jet classification using DL, we convert these detector measurements into multi-channel 2D images. Calorimeter and track information are transformed into image-like representations in the $(\eta, \phi)$ plane, with pixel intensities corresponding to energy deposits or track transverse momenta. Specifically, for the barrel region, HCAL tower data is upsampled to match the finer granularity of ECAL barrel crystals, and reconstructed track hits are placed into the image grid with their positions set by $(\eta, \phi)$ coordinates and intensities proportional to their transverse momentum. While the CMS calorimeter endcaps have different segmentation schemes from the barrel, for this study, we adopt the HCAL-centric image geometry, in which ECAL and track data are projected onto a grid matching the HCAL segmentation but at a finer barrel-like resolution~\cite{andrews_end--end_event}. This choice simplifies the image construction process for jet-centered views without significant information loss. The resulting full-detector image spans a $\eta$ range of $|\eta| < 3$ with a resolution of $\Delta \eta \times \Delta \phi = 280 \times 360$ pixels.

For each jet passing the selection criteria, a localized jet-view image window is extracted. The centroid of the reconstructed jet is determined, and the HCAL tower with the highest energy deposit within a $9 \times 9$ HCAL tower neighborhood is identified as the center. Around this center, a $125 \times 125$ pixel window is cropped to produce a jet image covering an approximate $\Delta R \lesssim 1$ region around the jet axis. Wrap-around padding is applied along the $\phi$ direction when jets are near the detector edges, ensuring seamless images. However, no padding is applied in the $\eta$ direction, effectively restricting the usable $\eta$ range to $|\eta| < 1.57$. This structured image generation pipeline transforms each selected jet into a high-resolution, three-channel image (ECAL, HCAL, and track), which serves as the input to our quark-gluon classification models. To gain qualitative insights into the spatial intensity patterns of the calorimeter and tracking data, we visualize the average per-pixel intensity distributions for gluon and quark jets across all detector channels (see Figure~\ref{fig:average_intensity}). Additionally, representative single-event images for both classes are presented in Figure~\ref{fig:single_event} to highlight individual variations and the heterogeneity of jet substructure.

\subsection{Data Preprocessing and Augmentation}
Effective data preprocessing and augmentation are essential for training robust DL models, particularly with specialized datasets like particle collision events. The raw pixel values in our dataset represent energy deposits in detector components. These values exhibit a wide dynamic range and skewed distributions, necessitating the use of a multi-stage preprocessing pipeline to normalize the data, mitigate outliers, and prepare the images for model input.

Initially, the images, originally sized at $125 \times 125$ pixels, undergo four main preprocessing steps. Each image has three channels corresponding to the ECAL, HCAL, and TRACK. First, \textit{Zero Suppression} removes low-energy noise by setting pixel values below $10^{-3}$ to zero. Let $I_{i,j,k}$ represent the original pixel value at spatial position $(i,j)$ for channel $k$, where $i$ and $j$ are the spatial indices and $k$ is the channel index ($k = 1$ for ECAL, $k = 2$ for HCAL, and $k = 3$ for TRACK). The modified pixel value, denoted $I'_{i,j,k}$, is defined by:
\begin{equation}
I'_{i,j,k} = \begin{cases} 
0 & \text{if } I_{i,j,k} < 10^{-3} \\
I_{i,j,k} & \text{otherwise} 
\end{cases}
\end{equation}
The next step is \textit{Global Z-Score Normalization}, which standardizes each channel independently. The mean ($\mu_k$) and standard deviation ($\sigma_k$) are computed globally across all images in the training dataset for each channel $k$. This process ensures each channel has an approximate mean of 0 and a standard deviation of 1. The normalized pixel value, $I''_{i,j,k}$, is calculated by:
\begin{equation}
I''_{i,j,k} = \frac{I'_{i,j,k} - \mu_k}{\sigma_k}
\end{equation}
Third, \textit{Outlier Clipping} mitigates extreme values by capping pixel intensities exceeding $500 \times \sigma_k$. The clipped pixel value, denoted $I'''_{i,j,k}$, is given by:
\begin{equation}
I'''_{i,j,k} = \min(I''_{i,j,k}, 500 \times \sigma_k)
\end{equation}
Finally, \textit{Sample-wise Min-Max Scaling} normalizes pixel values to the range $[0, 1]$ based on the minimum and maximum pixel values found across all three channels. The final preprocessed pixel value, $I_{\text{preproc}}(i,j,k)$, is calculated by:
\begin{equation}
I_{\text{preproc}}(i,j,k) = \frac{I'''_{i,j,k} - \min(I''')}{\max(I''') - \min(I''') + \epsilon}
\end{equation}
where $\epsilon$ is a small constant ($10^{-5}$) for numerical stability.

All input images were rescaled from the range $[0, 1]$ to $[0, 255]$ and converted to 8-bit unsigned integer format before augmentation. To standardize input dimensions and introduce scale and aspect ratio variation, \textit{Random Resized Cropping} was applied, selecting a random region within a scale range of $(0.8, 1.0)$ and an aspect ratio range of $(0.75, 1.33)$, subsequently resizing the crop to $224 \times 224$ pixels. \textit{Random Horizontal Flipping} is applied with a probability of 0.5, and \textit{Random Rotations} are applied by a randomly sampled angle $\theta$, uniformly selected from $[-20^\circ, 20^\circ]$:
\begin{equation}
I_{\text{flip}} = T_{\text{flip}}(I_{\text{PIL}})
\end{equation}
\begin{equation}
I_{\text{rotate}} = T_{\text{rotate}}(I_{\text{flip}}, \theta), \quad \theta \sim U(-20^\circ, 20^\circ)
\end{equation}

To introduce photometric variability during training, \textit{Random Color Jittering} was applied, adjusting image brightness ($\delta_b$), contrast ($\delta_c$), saturation ($\delta_s$) with random factors up to 0.2, and hue ($\delta_h$) by a factor of up to 0.1. Formally:
\begin{equation}
I_{\text{jitter}} = T_{\text{jitter}}(I_{\text{rrc}}, \delta_b, \delta_c, \delta_s, \delta_h)
\end{equation}

Following augmentation, images were rescaled from $[0, 255]$ to $[0, 1]$ and converted to floating-point tensors:
\begin{equation}
I_{\text{tensor}}(k,i,j) = \frac{I_{\text{jitter}}(i,j,k)}{255.0}
\end{equation}

For CNN architectures, including EfficientNet, CoAtNet, and ConvNeXt, inputs were standardized using ImageNet normalization statistics, with channel-wise means ($\mu_k$) of $[0.485, 0.456, 0.406]$ and standard deviations ($\sigma_k$) of $[0.229, 0.224, 0.225]$:
\begin{equation}
I_{\text{norm}}(k,i,j) = \frac{I_{\text{tensor}}(k,i,j) - \mu_k}{\sigma_k}
\end{equation}

For Transformer-based models, including ViT and Swin Transformers, this normalization step was omitted, and augmented tensors were used directly for training. Additionally, for ViT and Swin models, \textit{Mixup} augmentation was employed to improve generalization. Given two image-label pairs, $(I_a, y_a)$ and $(I_b, y_b)$, and a mixing coefficient $\lambda$ drawn from a Beta distribution $\text{Beta}(\alpha, \alpha)$ with $\alpha = 0.2$, new synthetic pairs were generated as:
\begin{align}
I_{\sim} &= \lambda I_a + (1 - \lambda) I_b \\
y_{\sim} &= \lambda y_a + (1 - \lambda) y_b
\end{align}

For validation, identical preprocessing was applied without any data augmentation. This included zero suppression, global Z-score normalization using training set statistics, outlier clipping, and sample-wise min-max scaling. The processed images $I_{\text{preproc}}$ were resized to $224 \times 224$ via bilinear interpolation, converted to tensors, and normalized according to the model-specific protocol. No augmentation transformations (e.g., random crops or flips) were applied during validation to ensure an unbiased performance assessment.

\subsection{Model Architectures}
This study evaluates several DL architectures for quark-gluon jet classification. We selected models from well-established CNN families and various Transformer-based approaches, including ensemble configurations. All models were primarily instantiated using the \texttt{timm} library~\cite{wightman_resnet_2021}. These models leveraged pre-trained weights, as detailed below. The aim of each architecture, \(M\), is to map a preprocessed input image 
\(\bm{x} \in \mathbb{R}^{H \times W \times C}\) (where \(H=224\), \(W=224\), \(C=3\)) to a probability distribution \(\bm{p} \in [0,1]^K\), where \(K=2\) classes.

Each standalone model \(m\) consists of a feature extraction backbone 
\(\Phi_m(\cdot; \bm{\theta}_{\Phi,m})\) and a classification head 
\(\Psi_m(\cdot; \bm{\theta}_{\Psi,m})\). The backbone 
\(\Phi_m\), with pre-trained weights \(\bm{\theta}_{\Phi,m}\), transforms 
\(\bm{x}\) into a \(D_m\)-dimensional feature vector \(\bm{f}_m\). We replace the original pre-trained classifier with a new linear layer 
\(\Psi_m\), which maps \(\bm{f}_m\) to \(K\)-dimensional logits 
\(\bm{z}_m\), as defined below:
\begin{equation}
\bm{z}_m = \bm{W}_m^T \bm{f}_m + \bm{b}_m
\label{eq:standalone_classifier_logits_v2}
\end{equation}
where \(\bm{W}_m \in \mathbb{R}^{D_m \times K}\) and \(\bm{b}_m \in \mathbb{R}^K\). Class probabilities \(\bm{p}_m\) are then computed as \(\bm{p}_m = \text{softmax}(\bm{z}_m)\).

\subsubsection{CNN Architectures} 
We considered several CNN families, all pre-trained on ImageNet-1k. These include ResNet50 (\(\Phi_{\text{RN50}}\))~\cite{fang_deep_2021}, known for its residual connections where each layer output is computed as \(y_l = H_l(x_l) + x_l\). EfficientNet-B0 (\(\Phi_{\text{EN-B0}}\))~\cite{tan_efficientnet_2019} uses MBConv blocks and compound scaling, where depth \(d = \alpha^\phi\), width \(w = \beta^\phi\), and resolution \(r = \gamma^\phi\) are scaled by a compound coefficient \(\phi\). RegNetY-002 (\(\Phi_{\text{RegY002}}\))~\cite{kiran_yenice_automated_2024} follows a quantized linear rule to determine block widths. ConvNeXt-Base (\(\Phi_{\text{CN-B}}\))~\cite{liu_convnet_2022} is a CNN adopting Transformer-inspired enhancements. For all CNN models, \(\bm{f}_m\) denotes the output feature obtained via a global average pooling (GAP) layer.

\subsubsection{Transformer Architectures} 
We also explored Transformer-based models~\cite{vaswani_attention_2017}, including ViT-Base/16 (\(\Phi_{\text{ViT-B}}\)) which processes 224×224 images split into 16×16 patches, pre-trained on ImageNet-1k. ViT divides the input image \(\bm{x}\) into \(N\) non-overlapping patches, expressed as \(\bm{x}_p = \{\bm{x}_p^1, \dots, \bm{x}_p^N\}\). Each patch is then embedded, positionally encoded, and prepended with a learnable class token. The sequence is processed through \(L\) Transformer encoder blocks, where each block applies multi-head self-attention (MHSA) and a multi-layer perceptron (MLP). Specifically, the intermediate output is computed as \(\bm{y}'_l = \text{MHSA}(\text{LN}(\bm{y}_{l-1})) + \bm{y}_{l-1}\), followed by \(\bm{y}_l = \text{MLP}(\text{LN}(\bm{y}'_l)) + \bm{y}'_l\). The final class token embedding after all encoder blocks becomes the image representation \(\bm{f}_{\text{ViT-B}}\). Additional Transformer-based models included Swin-Base (\(\Phi_{\text{Swin-B}}\))~\cite{liu_swin_2021}, CoAtNet-1 (\(\Phi_{\text{CoAt1}}\))~\cite{dai_coatnet_2021}, and MaxViT-Large (\(\Phi_{\text{MaxL}}\))~\cite{avidan_maxvit_2022}.

\subsubsection{Ensemble Models}
We constructed ensemble configurations by combining features from ViT-Base/16 pre-trained on ImageNet-21k (\(\Phi_{\text{ViT-21k}}\)) with a secondary backbone \(\Phi_X\). The features \(\bm{f}_{\text{ViT-21k}} = \Phi_{\text{ViT-21k}}(\bm{x})\) and \(\bm{f}_X = \Phi_X(\bm{x})\) were concatenated into a combined feature vector \(\bm{f}_{\text{ens}} = [\bm{f}_{\text{ViT-21k}} ; \bm{f}_X]\). This was fed into a two-layer MLP head \(\Psi_{\text{ens}}(\cdot; \bm{\theta}_{\Psi,\text{ens}})\), where the first hidden layer output is computed as \(\bm{h}_1 = \text{ReLU}(\bm{W}_{\text{ens,1}}^T \bm{f}_{\text{ens}} + \bm{b}_{\text{ens,1}})\), and the final logits as \(\bm{z}_{\text{ens}} = \bm{W}_{\text{ens,2}}^T \mathcal{D}(\bm{h}_1; \delta) + \bm{b}_{\text{ens,2}}\). Here, the hidden dimension is set to \(D_h = 512\), and \(\mathcal{D}(\cdot; \delta)\) represents dropout with a probability \(\delta = 0.1\). The backbones \(\Phi_X\) considered include EN-B0 (\(\Phi_{\text{EN-B0}}\)), RegY002 (\(\Phi_{\text{RegY002}}\)), CN-B (\(\Phi_{\text{CN-B}}\)), MaxViT-Base (\(\Phi_{\text{MaxB}}\)), Swin-Base trained on ImageNet-21k (\(\Phi_{\text{Swin-21k}}\)), and CoAtNet-0 (\(\Phi_{\text{CoAt0}}\)).

Additionally, we evaluated a triple-model ensemble combining ViT-21k, CN-B, and Swin-Base trained on ImageNet-1k (\(\Phi_{\text{Swin-B}}\)). In this case, the concatenated feature vector is given by \(\bm{f}_{\text{ens3}} = [\bm{f}_{\text{ViT-21k}} ; \bm{f}_{\text{CN-B}} ; \bm{f}_{\text{Swin-B}}]\) and is passed to a similar two-layer MLP head \(\Psi_{\text{ens3}}\). All parameters, including those of the backbones (\(\bm{\theta}_{\Phi,m}\)) and the MLP heads (\(\bm{\theta}_{\Psi,m}\), \(\bm{\theta}_{\Psi,\text{ens}}\), and \(\bm{\theta}_{\Psi,\text{ens3}}\)) were jointly trained (see Sec.~\ref{3.4}).

\subsection{Experimental Setup and Training Procedure}\label{3.4}
For all models in this study, we employed a consistent experimental configuration to ensure fair comparisons. All experiments were conducted on an NVIDIA Tesla P100 GPU (16GB VRAM), 2 CPU cores (Intel Xeon), and 13GB RAM. We used the AdamW optimizer with an initial learning rate of $1\times10^{-4}$ for the classification head and a weight decay of $1\times10^{-4}$. A cosine annealing learning rate scheduler was applied with a maximum of 50 iterations. Gradual unfreezing was employed: at epoch 5, block 3 of the Transformer encoder was unfrozen; at epoch 8, blocks 2 and 3 were unfrozen. Unfrozen parameters in these blocks were added to the optimizer with a learning rate of $1\times10^{-6}$. Models were trained for a maximum of 20 epochs with early stopping based on validation loss, applying a patience of 5 epochs. A batch size of 32 was used for both training and validation. Mixed precision training was enabled using PyTorch's `GradScaler' and `autocast' for accelerated computation on GPUs. Performance was evaluated using accuracy, precision, recall, F1-score, ROC-AUC, and confusion matrices on the validation set. Additionally, inference time per image was measured by averaging the forward pass time for a single randomly generated $224\times224\times3$ image sample. To account for experimental variability, each experiment was independently repeated three times using different random seeds. The final reported results correspond to the mean $\pm$ standard deviation across these three runs. The total number of trainable parameters was recorded at each stage. Training time per epoch and total training duration were logged for runtime analysis.

\begin{table}[ht]
\centering
\caption{Performance comparison of benchmarked models on 7k samples from the quark-gluon dataset. Results are reported as \textit{mean $\pm$ standard deviation} over three runs with different random seeds. \textbf{Bold} indicates the best performance, while \ul{underline} marks the second-best result.}
\label{tab:results}
\begin{sc}
\resizebox{\columnwidth}{!}{%
\begin{tabular}{lccccc|ccc}
\toprule[1.5pt]
\textbf{Model} & \textbf{Accuracy (\%) (\textcolor{mygreen}{$\uparrow$})} & \textbf{Precision (\%) (\textcolor{mygreen}{$\uparrow$})} & \textbf{Recall (\%) (\textcolor{mygreen}{$\uparrow$})} & \textbf{F1 Score (\%) (\textcolor{mygreen}{$\uparrow$})} & \textbf{ROC-AUC (\%) (\textcolor{mygreen}{$\uparrow$})} & \textbf{\# Params (\textcolor{myred}{$\downarrow$})} & \textbf{Train Time (\textcolor{myred}{$\downarrow$})} & \textbf{Inference Time (ms) (\textcolor{myred}{$\downarrow$})} \\
\midrule[1pt]
ViT + MaxViT & \ul{70.29 $\pm$ 0.0224} & \textbf{77.35 $\pm$ 0.0397} & 76.45 $\pm$ 0.0613 & \textbf{72.02 $\pm$ 0.0392} & \textbf{76.65 $\pm$ 0.0287} & 236M & 54m 41s & 276.11 \\
ViT + ConvNeXt & \textbf{70.57 $\pm$ 0.0354} & 72.67 $\pm$ 0.0477 & 75.47 $\pm$ 0.0914 & \ul{71.33 $\pm$ 0.0308} & \ul{76.25 $\pm$ 0.0304} & 287M & 34m 27s & 354.33 \\
ViT + EfficientNet & 70.00 $\pm$ 0.0186 & 71.26 $\pm$ 0.0320 & 76.36 $\pm$ 0.0584 & 70.75 $\pm$ 0.0241 & 76.14 $\pm$ 0.0229 & 190.8M & 17m 8s & 67.96 \\
RegNetY & 69.43 $\pm$ 0.0164 & 71.30 $\pm$ 0.0310 & 66.05 $\pm$ 0.0508 & 68.58 $\pm$ 0.0200 & 75.89 $\pm$ 0.0224 & 2.98M & 10.41m & 10.89 \\
ViT + Swin & 69.86 $\pm$ 0.0235 & \ul{74.43 $\pm$ 0.0417} & 80.93 $\pm$ 0.0752 & 71.27 $\pm$ 0.0380 & 75.62 $\pm$ 0.0213 & 183M & 27m 32s & 44.74 \\
ViT + RegNetY & 69.79 $\pm$ 0.0148 & 70.54 $\pm$ 0.0343 & 69.85 $\pm$ 0.0616 & 70.19 $\pm$ 0.0227 & 74.92 $\pm$ 0.0148 & 89.77M & 24.07m & 169.11 \\
ConvNeXt & 67.57 $\pm$ 0.0241 & 72.93 $\pm$ 0.0308 & 75.24 $\pm$ 0.0597 & 70.33 $\pm$ 0.0401 & 72.91 $\pm$ 0.0276 & 89M & 8m 50s & 54.20 \\
ViT + CoAtNet & 66.79 $\pm$ 0.0113 & 67.59 $\pm$ 0.0130 & 67.87 $\pm$ 0.0402 & 67.73 $\pm$ 0.0194 & 71.79 $\pm$ 0.0108 & 0.79M & 7m 52s & 20.21 \\
ViT + ConvNeXt + Swin & 66.64 $\pm$ 0.0280 & 68.01 $\pm$ 0.0275 & 78.32 $\pm$ 0.0492 & 69.09 $\pm$ 0.0314 & 71.11 $\pm$ 0.0159 & 312M & 17m 45s & 92.66 \\
ViT & 69.29 $\pm$ 0.0416 & 69.34 $\pm$ 0.0474 & 73.68 $\pm$ 0.0448 & 70.09 $\pm$ 0.0150 & 69.28 $\pm$ 0.0419 & 85M & 11.1m & 55.52 \\
Swin & 69.29 $\pm$ 0.0555 & 69.36 $\pm$ 0.0547 & \ul{84.92 $\pm$ 0.0559} & 70.93 $\pm$ 0.0390 & 69.28 $\pm$ 0.0557 & 87M & 23.3m & 36.58 \\
CoAtNet & 61.29 $\pm$ 0.0238 & 66.83 $\pm$ 0.0416 & \textbf{88.00 $\pm$ 0.1214} & 67.26 $\pm$ 0.0619 & 66.65 $\pm$ 0.0285 & 82M & 15m 30s & 68.50 \\
MaxViT & 66.36 $\pm$ 0.0152 & 65.69 $\pm$ 0.0186 & 78.95 $\pm$ 0.0549 & 69.29 $\pm$ 0.0210 & 66.34 $\pm$ 0.0153 & 119M & 59.3m & 141.63 \\
ResNet & 63.79 $\pm$ 0.0146 & 63.13 $\pm$ 0.0134 & 74.82 $\pm$ 0.0723 & 66.97 $\pm$ 0.0365 & 63.77 $\pm$ 0.0145 & 15M & 5m 30s & 103.15 \\
EfficientNet & 59.57 $\pm$ 0.0205 & 60.33 $\pm$ 0.0225 & 57.33 $\pm$ 0.0361 & 58.57 $\pm$ 0.0252 & 59.58 $\pm$ 0.0205 & 29M & 6.1m & 58.29 \\
\bottomrule[1.5pt]
\end{tabular}
}
\end{sc}
\end{table}

\section{Results and Discussion}
\label{results_discussion}

\subsection{Performance Comparisons}
Table~\ref{tab:results} reports a comprehensive comparison of model performance. ViT-based architectures and their hybrid variants consistently outperform pure CNN models in quark-gluon jet classification across most metrics. Notably, \textit{ViT + MaxViT} achieved the highest F1 score (72.02\%) and ROC-AUC (76.65\%), while \textit{ViT + ConvNeXt} obtained the top accuracy (70.57\%). These results suggest that the ability of ViTs to capture global context provides a tangible advantage in this task, where long-range feature dependencies are likely relevant. Hybrid combinations generally improved performance compared to standalone models. For example, adding \textit{MaxViT} or \textit{ConvNeXt} to ViT results in noticeable gains in F1 score and ROC-AUC over either component alone. This indicates that combining Transformers' global attention with CNNs' inductive biases toward locality and translation equivariance leads to more effective feature representations for jet classification. Among standalone models, the \textit{Swin Transformer} delivered strong results (recall: 84.92\%, F1: 70.93\%), outperforming most CNN baselines. Interestingly, while \textit{CoAtNet} achieved the highest recall (88.00\%), its precision and overall F1 score remained lower, highlighting a tendency to over-predict one class. This tradeoff between recall and precision is important in physics applications where controlling false positives may be as critical as maximizing sensitivity. Traditional CNNs such as \textit{ResNet} and \textit{EfficientNet} underperformed on primary classification metrics. \textit{ResNet}, despite faster training, lagged in F1 score (66.97\%) and ROC-AUC (63.77\%), suggesting that CNNs' locality-constrained receptive fields limit their capacity to capture the complex, non-local structures present in jet images.

These findings point to a consistent pattern: architectures incorporating global attention mechanisms outperform those relying solely on local convolutions. The improved ability to model long-range dependencies appears to be crucial for this classification task, likely due to the extended spatial correlations present in quark and gluon jet images. This trend highlights a shift in the preferred architecture class for HEP applications, with Transformer-based models now setting the baseline.

\begin{figure}[!ht] 
\centering
\subfloat[Effect of dataset size on model performance.\label{fig:dataset_size}]{
   \includegraphics[width=0.3\linewidth]{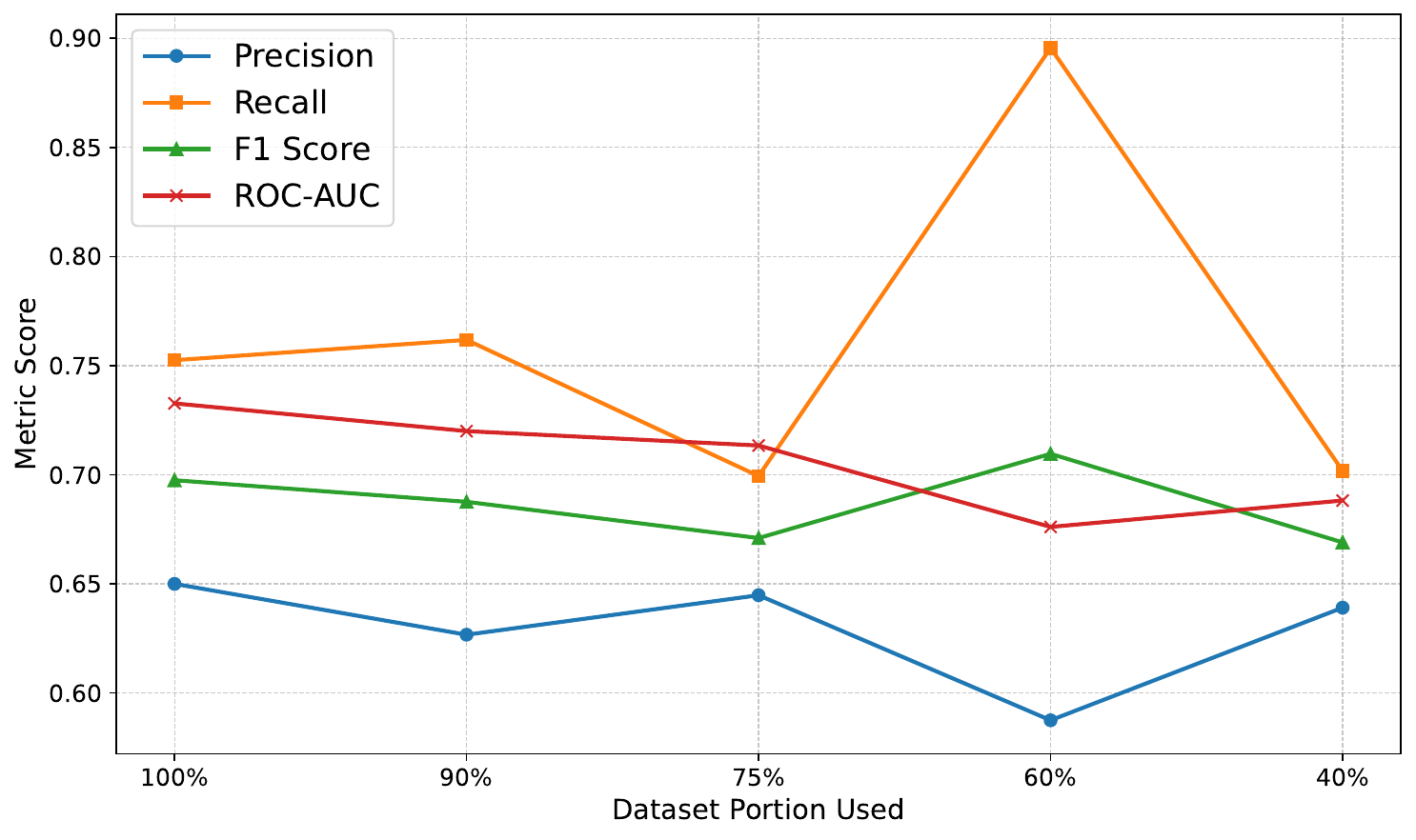}}
\hfill
\subfloat[Impact of ViT model variant size on performance.\label{fig:vit_variant}]{
    \includegraphics[width=0.3\linewidth]{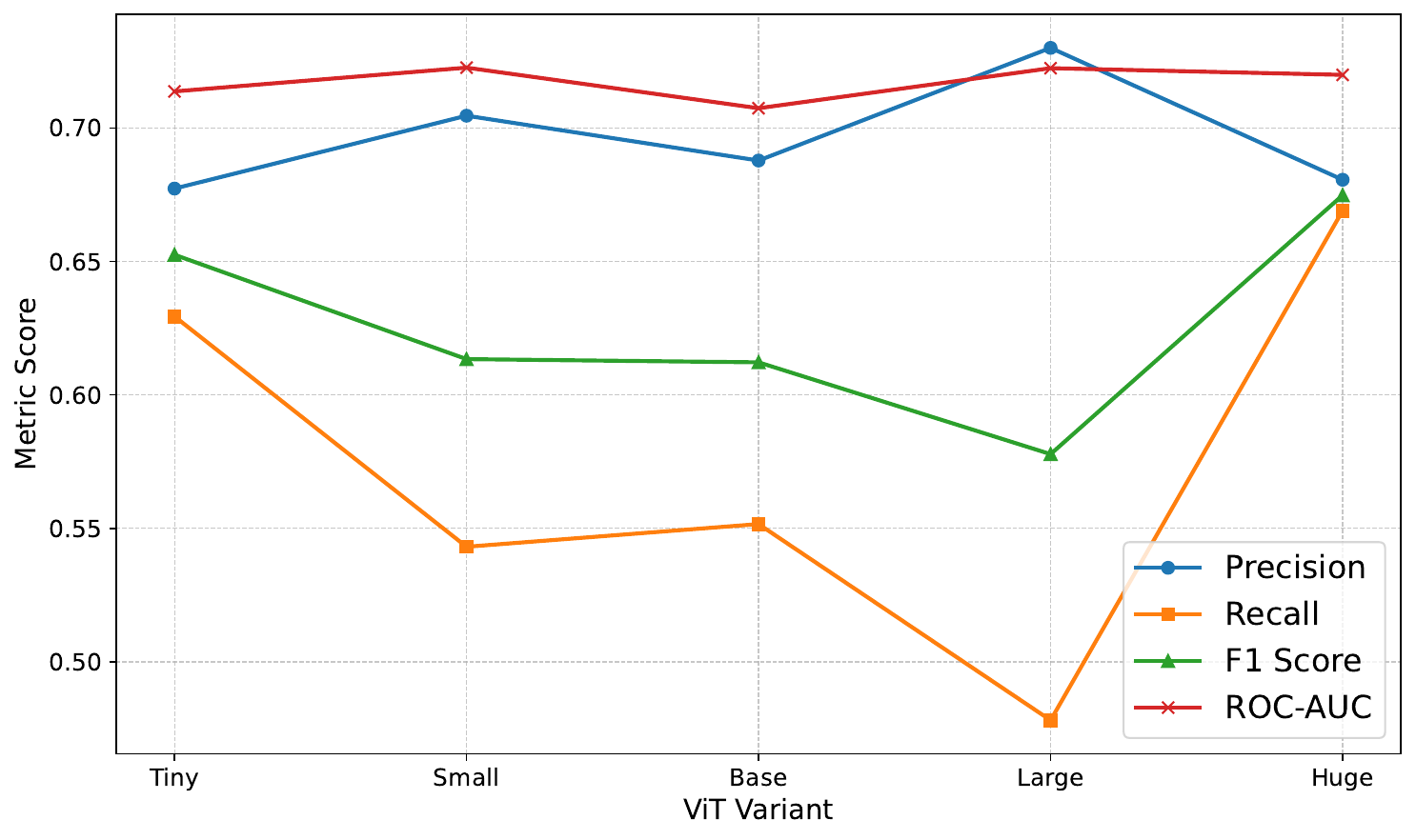}}
\hfill
\subfloat[Effect of batch size on performance.\label{fig:batch_size}]{
    \includegraphics[width=0.3\linewidth]{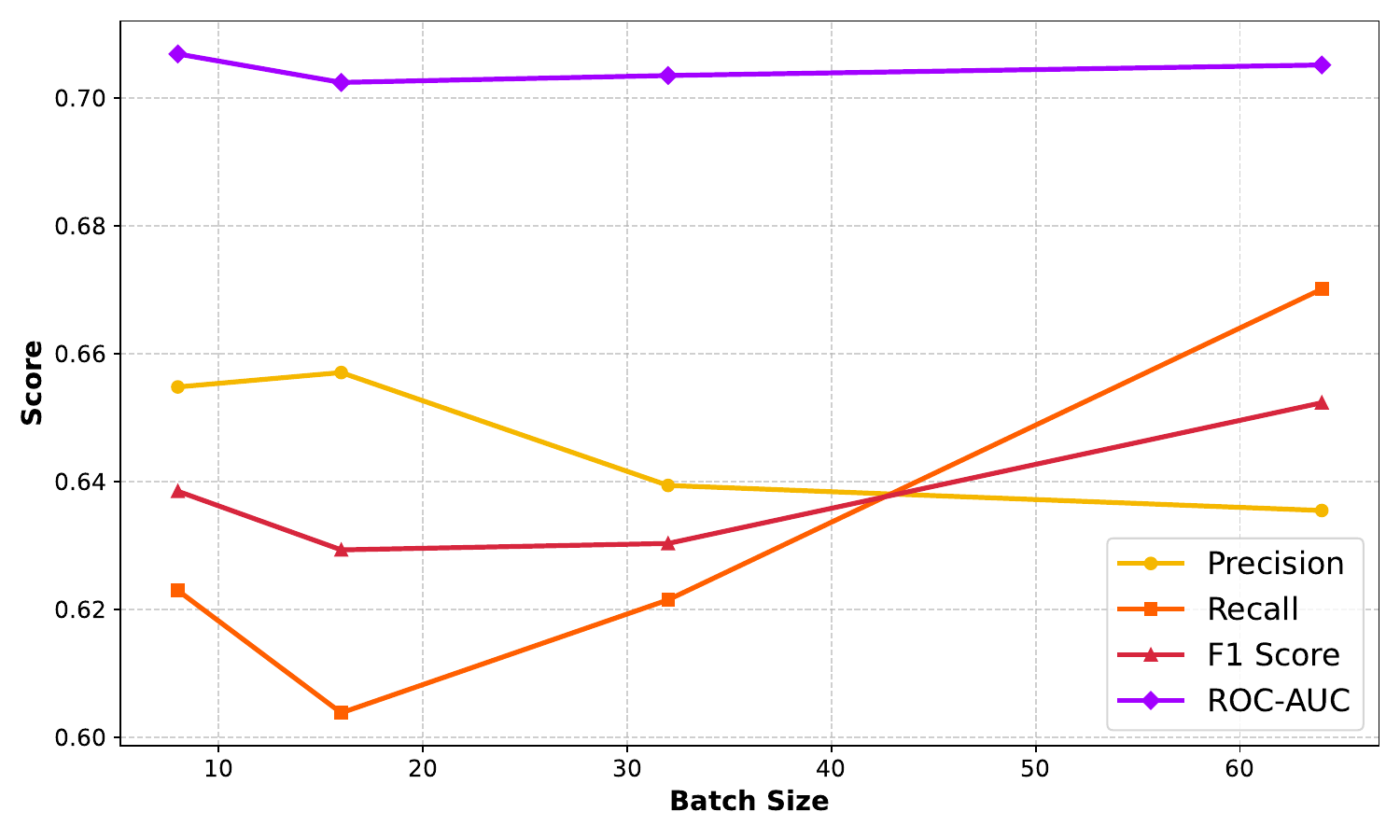}}
\hfill
\subfloat[Influence of learning rate on model accuracy.\label{fig:learning_rate}]{
    \includegraphics[width=0.3\linewidth]{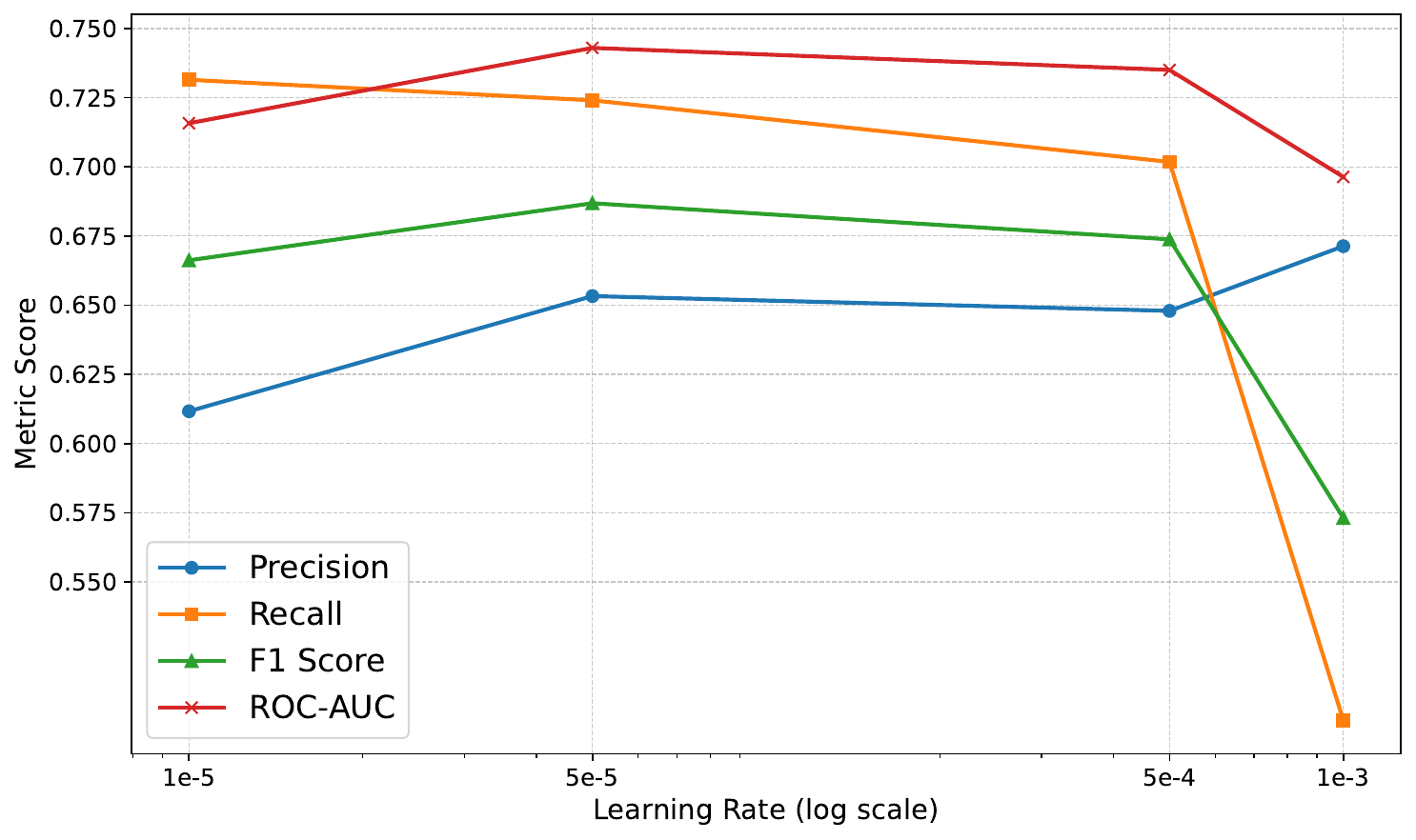}}
\hfill
\subfloat[Comparison of optimizers and their effect on performance.\label{fig:optimizers}]{
    \includegraphics[width=0.3\linewidth]{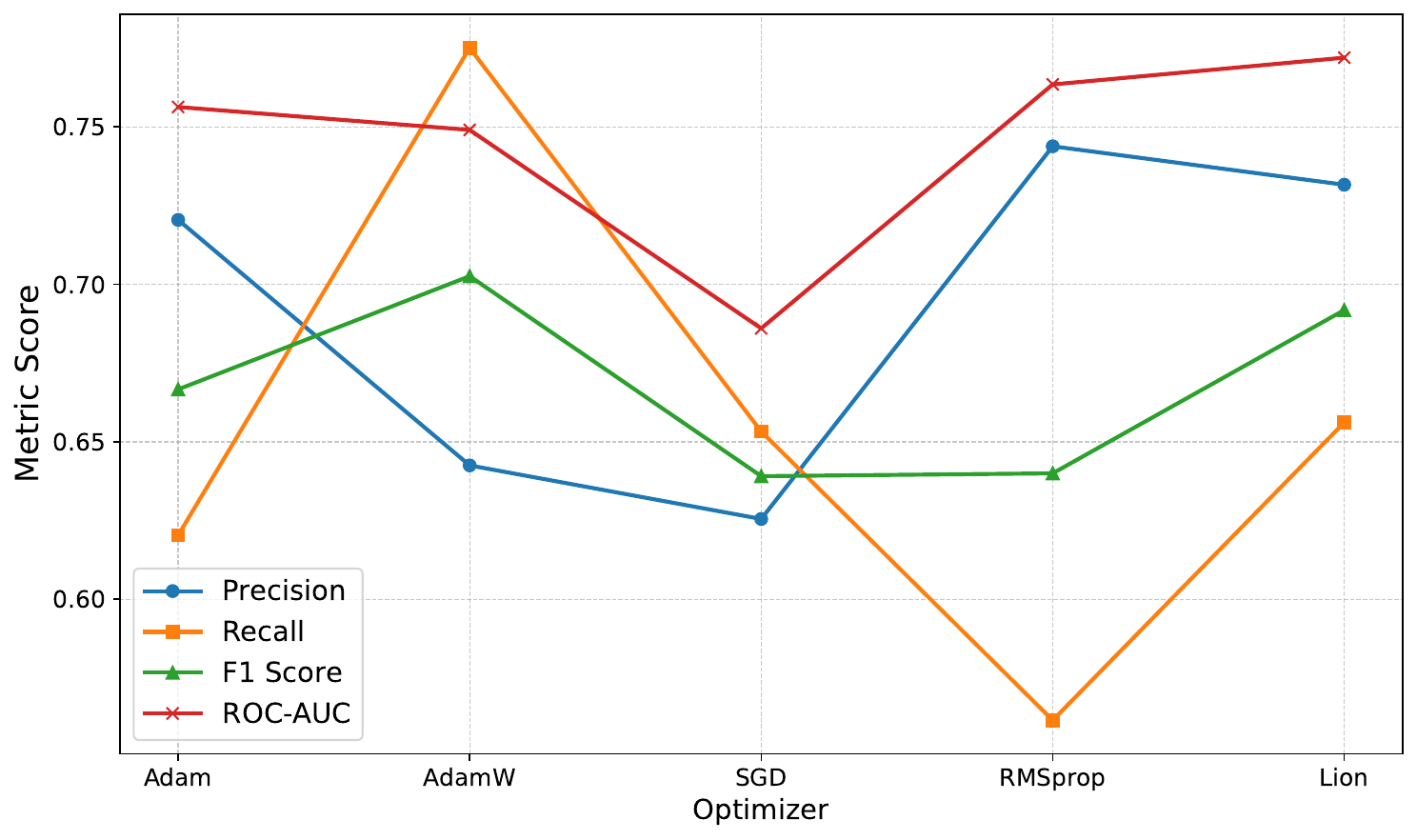}}
\hfill
\subfloat[Impact of weight decay regularization.\label{fig:weight_decay}]{
    \includegraphics[width=0.3\linewidth]{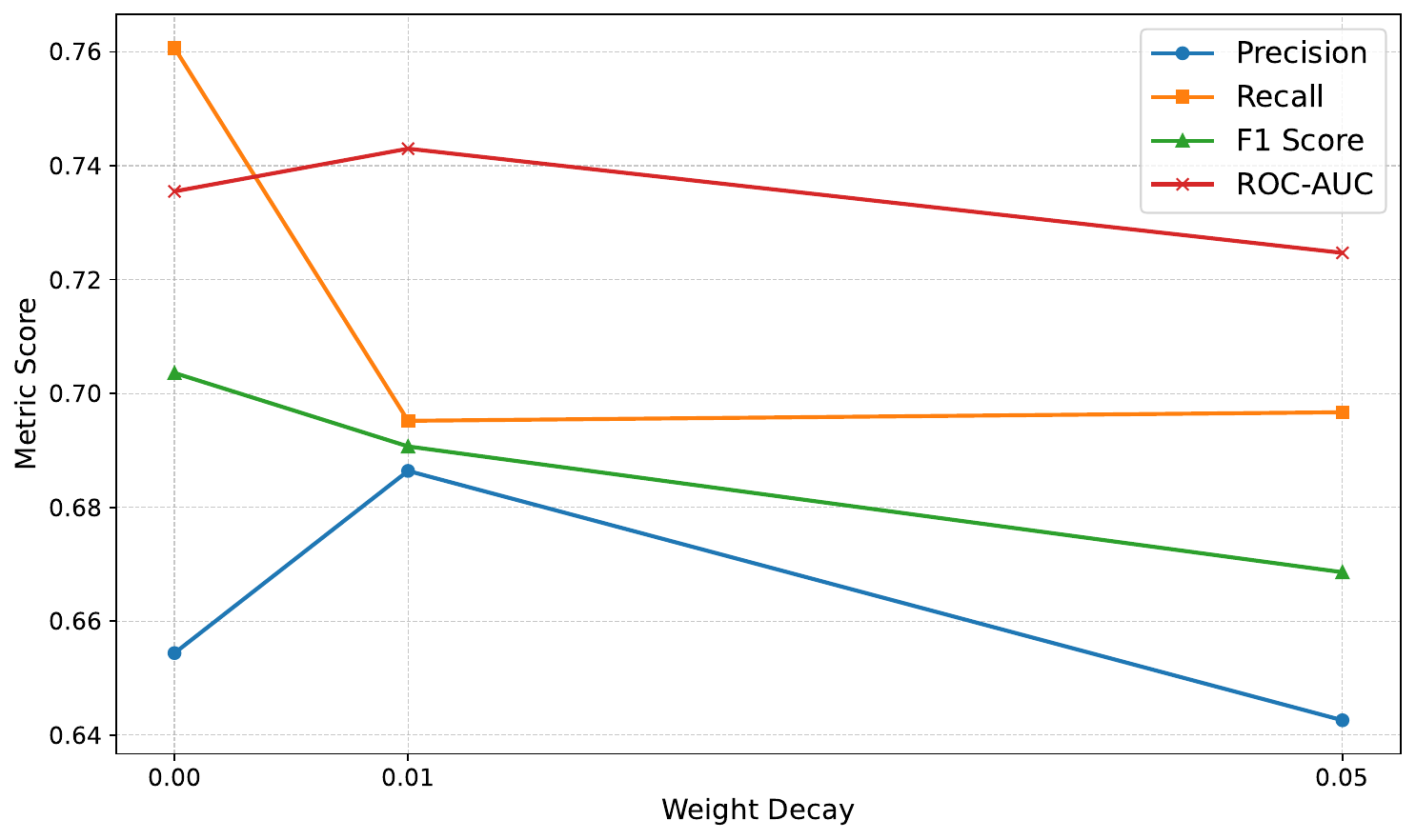}}
\vspace{0.3em}
\centerline{
\subfloat[Performance trends across training epochs.\label{fig:epochs}]{
    \includegraphics[width=0.3\linewidth]{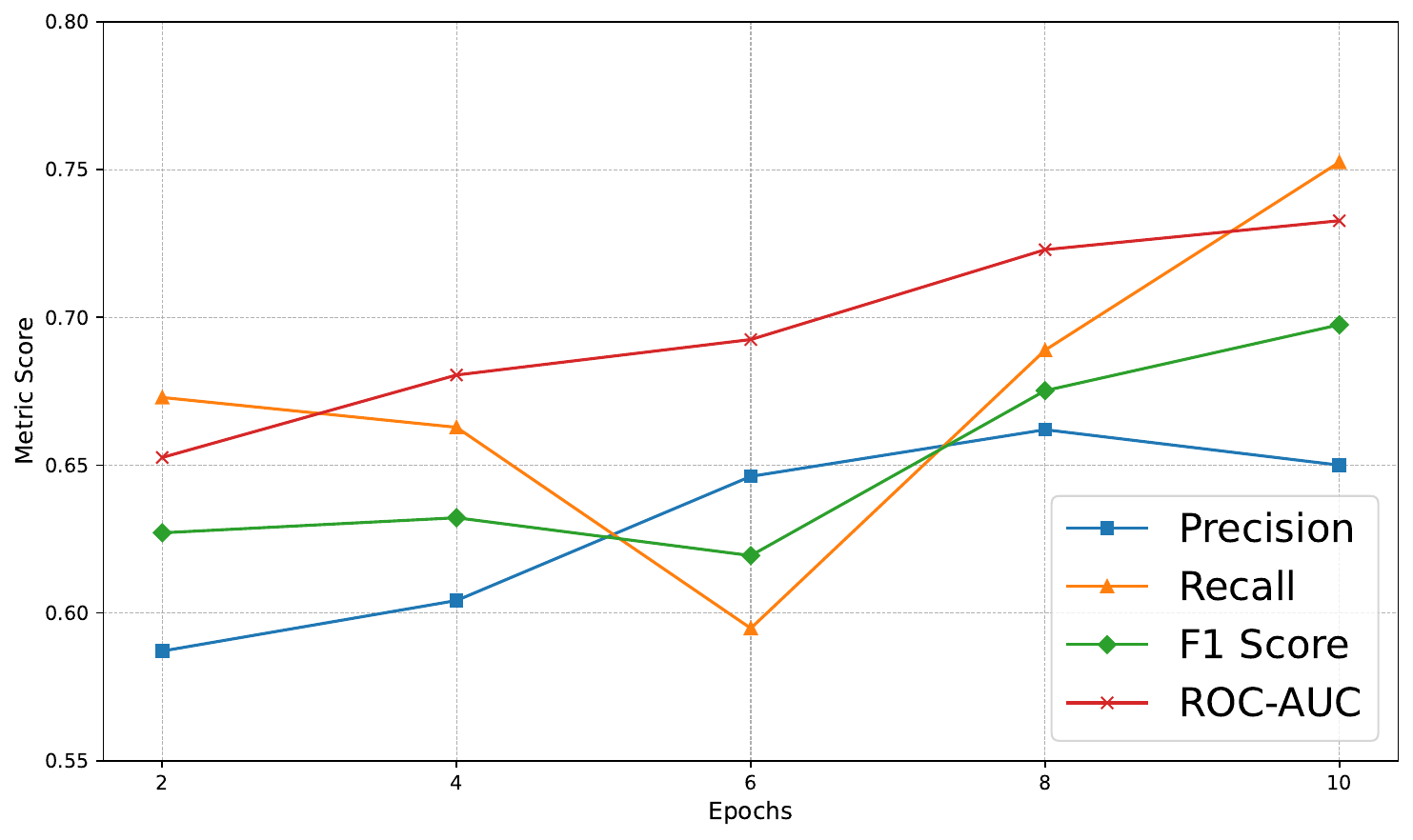}}
\hspace{1em}
\subfloat[Effect of dropout rate on final performance.\label{fig:dropout}]{
    \includegraphics[width=0.3\linewidth]{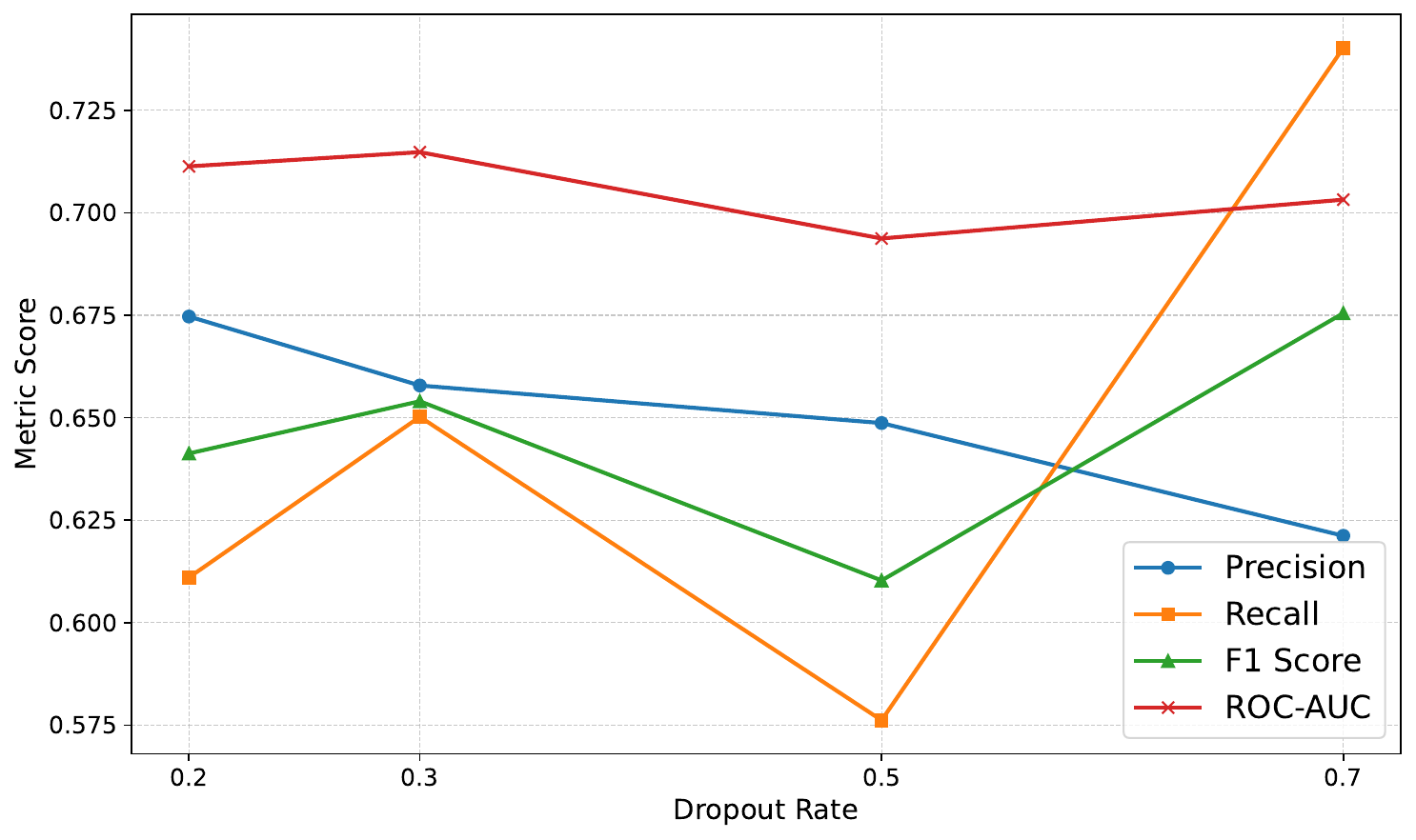}}
}
\caption{Ablation study summarizing the effect of key training and architectural hyperparameters on model performance. Each subfigure isolates a single factor while holding others constant, illustrating its individual impact.}
\label{fig:ablation}
\end{figure}

\subsection{Sensitivity Analysis}
To evaluate the robustness and stability of our proposed model, we performed an extensive sensitivity analysis by systematically varying key hyperparameters and architectural configurations, as illustrated in Figure~\ref{fig:ablation}. Starting with dataset size (Figure~\ref{fig:dataset_size}), the model exhibited impressive resilience, maintaining a respectable F1 score (0.7096) and an elevated recall (0.8956) even when trained on only 60\% of the data. Next, when comparing different transformer model variants (Figure~\ref{fig:vit_variant}), the ViT-Huge model achieved the strongest recall-F1 balance, albeit with higher computational demands, confirming the advantage of larger architectures for complex medical imaging tasks. Regarding batch size (Figure~\ref{fig:batch_size}), a batch size of 64 yielded the highest recall and F1 score, while a batch size of 8 offered a marginally better ROC-AUC. In the learning rate comparison (Figure~\ref{fig:learning_rate}), $5\times10^{-5}$ consistently provided the best trade-off across metrics, achieving higher precision, recall, and ROC-AUC while avoiding the instability associated with more aggressive rates like $1\times10^{-3}$. Among optimizers (Figure~\ref{fig:optimizers}), Lion emerged as the top performer, delivering the highest ROC-AUC (0.7720) and competitive F1 scores, outperforming Adam, AdamW, and RMSprop, in line with its growing success in vision-based applications. For regularization (Figure~\ref{fig:weight_decay}), a weight decay value of 0.01 struck an optimal balance by enhancing generalization without excessively restricting model capacity. Performance trends across training epochs (Figure~\ref{fig:epochs}) revealed steady improvements in validation accuracy, F1 score, and ROC-AUC, with marked gains after unfreezing the ViT backbone at epoch 6 and the MaxViT backbone at epoch 10. The final epoch yielded the highest F1 score (0.6975) and ROC-AUC (0.7327), highlighting the benefit of staged fine-tuning. Lastly, in assessing dropout rates (Figure~\ref{fig:dropout}), a value of 0.3 proved to be the most reliable, offering balanced improvements, while higher rates, such as 0.,7 favored recall at the expense of precision.

\section{Conclusion}
This study systematically evaluated ViT and ViT-CNN hybrid architectures for end-to-end quark-gluon jet classification, leveraging multi-channel calorimeter images from publicly available CERN (CMS) data. Our findings robustly demonstrate that ViT-based approaches, particularly hybrids like \textit{ViT+MaxViT} and \textit{ViT+ConvNeXt}, significantly outperform established CNN baselines in key metrics such as F1-score, ROC-AUC, and accuracy. ViTs excel because they can analyze the entire jet image at once, helping them spot subtle, widespread patterns that are key to telling these jets apart, even amidst experimental noise. Our work provides the first systematic comparison of ViTs for this specific task using public data, offering strong performance results and a ready-to-use dataset. This helps make advanced image analysis tools more accessible for particle physics research.

\paragraph{Limitations}
Key limitations of this study include: (i) The results are based on simulated data; further tests on real, more recent experimental data with different conditions are needed; (ii) These advanced ViT models are powerful but require significant computing resources, which could be a challenge for real-time applications without further optimization; (iii) While we know these models work well, more research is needed to fully understand, from a physics standpoint, exactly what features they are learning from the images.

% References Section

{
\tiny
\bibliographystyle{splncs04}
\bibliography{main}
}

%%%%%%%%%%%%%%%%%%%%%%%%%%%%%%%%%%%%%%%%%%%%%%%%%%%%%%%%%%%%

% \appendix

% \section{Additional Details of Related Work}
% \label{app:related_work}

%%%%%%%%%%%%%%%%%%%%%%%%%%%%%%%%%%%%%%%%%%%%%%%%%%%%%%%%%%%%

\end{document}